# Multi-Scenario Combination Based on Multi-Agent Reinforcement Learning to Optimize the Advertising Recommendation System

Yang Zhao[1]
Columbia University
New York, USA
yangzhaozyang@gmail.com

Chang Zhou[2,*]
Columbia University
New York, USA
* Corresponding author: mmchang042929@gmail.com

Jin Cao[2]
Independent Researcher
Dallas, USA
caojinscholar@gmail.com

Yi Zhao[3]
Independent Researcher
Sunnyvale, USA
zhaoyizjuee@gmail.com

Shaobo Liu[4]
Independent Researcher
Broomfield, USA
shaobo1992@gmail.com

Chiyu Cheng[5]
University of California, Irvine
Seattle, USA
cypersonal6@gmail.com

Xingchen Li[5]
University of Southern California
Los Angeles, USA
stellali0919@gmail.com

*Abstract*—This paper explores multi-scenario optimization on large platforms using multi-agent reinforcement learning (MARL). We address this by treating scenarios like search, recommendation, and advertising as a cooperative, partially observable multi-agent decision problem. We introduce the Multi-Agent Recurrent Deterministic Policy Gradient (MA-RDPG) algorithm, which aligns different scenarios under a shared objective and allows for strategy communication to boost overall performance. Our results show marked improvements in metrics such as click-through rate (CTR), conversion rate, and total sales, confirming our method's efficacy in practical settings.

## I. INTRODUCTION

In the dynamic landscape of e-commerce, platforms host a multitude of interconnected scenarios including search, recommendation, and advertisements. Each of these sub-scenarios further divides into specialized categories, default ranking and store search within the search scenario. Current optimization techniques typically treat these scenarios in isolation, leading to disjointed user experiences and suboptimal overall performance. Users frequently navigate between different scenarios, and independent optimization can result in conflicts and inefficiencies. To address these issues, we explores using Multi-Agent Reinforcement Learning to achieve a cohesive and enhanced user experience across multiple scenarios. We develop a system where each scenario's strategy considers the overall platform performance. This paper investigates the application of the MA-RDPG algorithm, examining its impact on key performance indicators (KPIs) and overall platform efficiency.

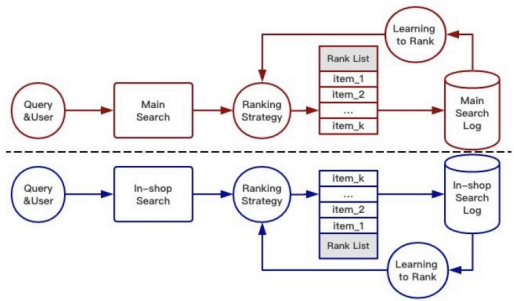

Figure 1.  Two search engines independently optimized.

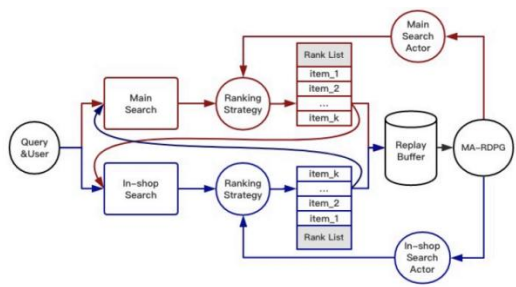

Figure 2.  Using MA-RDPG, 2 search engines jointly optimized.

## II. THEORETICAL FRAMEWORK

The theoretical foundation of this research is based on the principles of reinforcement learning (RL) and multi-agent systems. The core concept involves agents interacting with an environment to maximize cumulative rewards. In a multi-agent setup, agents can be cooperative, competitive, or mixed. For our study, we adopt a fully cooperative model where all agents share the same objective function, aiming to optimize the global performance of the platform.

### A. Deep Recurrent Q-Networks (DRQN)

Traditional RL approaches assume fully observable environments. However, in practical applications like e-commerce, the environment is partially observable. Deep Recurrent Q-Networks (DRQN) address this by utilizing recurrent neural networks (RNNs) to encode historical observations, thereby maintaining a memory of past states and actions.

### B. Actor-Critic Methods

Our approach is inspired by the Deterministic Policy Gradient (DPG) method, specifically the Deep Deterministic Policy Gradient (DDPG), which combines the strengths of Q-learning and policy gradient methods. The actor-critic framework, comprising a central critic that evaluates actions and multiple actors that propose actions, serves as the basis for our model.

## III. LITERATURE REVIEW

Reinforcement learning (RL)[2] has advanced significantly and is utilized in areas such as natural language processing[3-6], computer vision[7-12], deep learning[13-16], and machine learning[17-22]. Building on foundational research in RL and multi-agent systems, Learning to Rank (L2R) algorithms have evolved in online systems from point-wise to list-wise methods. In multi-agent RL (MARL), there have been substantial developments in both cooperative and competitive dynamics, applied in fields from robotics to resource management.

However, the use of MARL in e-commerce to optimize interrelated scenarios is still underexplored. Our study addresses this by proposing a unified framework that promotes scenario cooperation, aiming to prevent conflicts and enhance overall performance. The findings of Li et al.[1] demonstrated that combining data from two distinct components significantly improves model performance. This inspired us to consider integrating inputs from various sources in our current research.

## IV. METHODOLOGY

The methodology tackles the multi-scenario optimization in e-commerce as a cooperative, partially observable multi-agent decision problem using the MA-RDPG algorithm, which merges DRQN and DPG. This allows agents to recall past interactions and optimize actions in continuous spaces, enhancing robust predictions. We follow Li et al.[1] approach, utilizing multiple rounds of majority voting to ensure reliable results with a limited dataset.

### A. Data Collection and Preprocessing

We gather data from an unpublished e-commerce dataset, including search queries, click data, and purchase histories. This data is preprocessed to generate training samples for the reinforcement learning model.

### B. Algorithm Implementation

The MA-RDPG is implemented with a global critic for performance evaluation, scenario-specific actors for action generation, and an LSTM-based communication module to facilitate information sharing. The critic estimates the Q-value function, while actors decide on actions based on the current state and inter-agent communications.

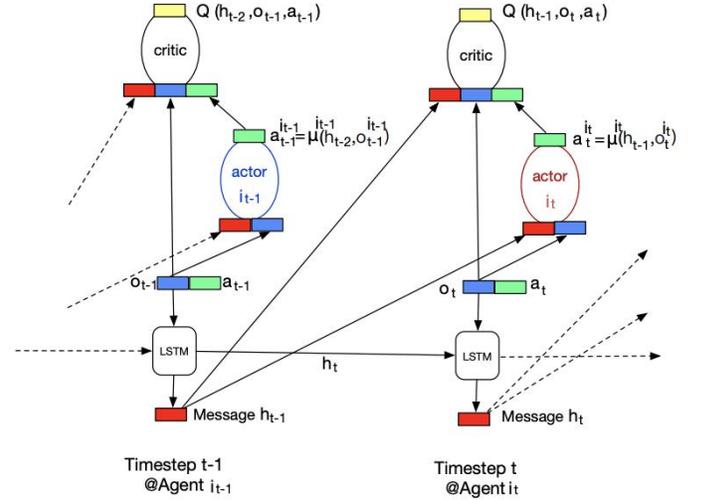

Figure 3. Detailed structure of the MA-RDPG algorithm. ( *The central referee simulates the "action-value" function Q(ht−1, ot, at), which represents the overall benefit of taking action at when receiving information ht−1 and observation ot.* )

### C. Training and Evaluation

The training process involves continuous interaction with the environment, using a replay buffer to store experiences and update the model via mini-batch gradient descent. The performance of the algorithm is evaluated using standard A/B testing, comparing key metrics such as CTR, conversion rate, and gross merchandise volume (GMV) against baseline models.

We train the centralized critic network Q using the Bellman formula as in Q-learning. We minimize the following loss function:

$$L(\phi) = E_{h_{t-1}, o_t}[(Q(h_{t-1}, o_t, a_t; \phi) - y_t)^2] \quad (1)$$

The update of the private actor network is based on maximizing the overall expectation. Assuming that at time t, Ait is active, then the objective function is:

$$J(\theta^{i_t}) = E_{h_{t-1}, o_t}[Q(h_{t-1}, o_t, a; \phi)|_{a=\mu^{i_t}(h_{t-1}, o_t; \theta^{i_t})}] \quad (2)$$

According to the chain rule, the goal of module training is to minimize the following function:

$$L(\psi)$$
$$= E_{h_{t-1},o_t}[(Q(h_{t-1},o_t,a_t;\phi) - y_t)^2 |_{h_{t-1}=LSTM(h_{t-2},[o_{t-1};a_{t-1}];\psi)}]$$
$$- E_{h_{t-1},o_t}[Q(h_{t-1},o_t,a_t;\phi)|_{h_{t-1}=LSTM(h_{t-2},[o_{t-1};a_{t-1}];\psi)}] \quad (3)$$

The training process, detailed in Table 1, involves using a replay buffer to store interactions between agents and the environment, which are then updated in minibatches. During each training session, we select and train several minibatches in parallel, simultaneously updating the actor and critic networks.

---

### ALGORITHM 1: MA-RDPG

**Input:** The environment
**Output:** $\theta = \{\theta^1, \ldots, \theta^N\}$
Initialize the parameters $\theta = \{\theta^1, \ldots, \theta^N\}$ for the $N$ actor networks and $\phi$ for the centralized critic network;
Initialized the replay buffer $R$;
**foreach** training step $e$ **do**
    **for** $i = 1$ to $M$ **do**
        $h_0$ = initial message, $t = 1$;
        **while** $t < T$ and $o_t \ne terminal$ **do**
            Select the action $a_t = \mu^{i_t}(h_{t-1}, o_t)$ for the active agent $i_t$;
            Receive reward $r_t$ and the new observation $o_{t+1}$;
            Generate the message $h_t = LSTM(h_{t-1}, [o_t; a_t])$;
            $t = t + 1$;
        **end**
        Store episode $\{h_0, o_1, a_1, r_1, h_1, o_2, r_2, h_3, o_3, \ldots\}$ in $R$;
    **end**
    Sample a random minibatch of episodes $B$ from replay buffer $R$;
    **foreach** episode in $B$ **do**
        **for** $t = T$ downto $1$ **do**
            Update the critic by minimizing the loss:
            $L(\phi) = (Q(h_{t-1}, o_t, a_t; \phi) - y_t)^2$, where
            $y_t = r_t + \gamma Q(h_t, o_{t+1}, \mu^{i_{t+1}}(h_t, o_{t+1}); \phi)$;
            Update the $i_t$-th actor by maximizing the critic;
            $J(\theta^{i_t}) = Q(h_{t-1}, o_t, a; \phi)|_{a=\mu^{i_t}(h_{t-1}, o_t; \theta^{i_t})}$;
            Update the communication component;
        **end**
    **end**
**end**

---

## V. RESULTS

### A. Experiment Analysis

The results indicate that our MA-RDPG algorithm outperforms existing algorithms, particularly the L2R+L2R algorithm commonly used by e-commerce platforms. While L2R+L2R focuses solely on individual scenario optimization, MA-RDPG enhances overall benefits by fostering cooperative interactions between scenarios, which significantly improves GMV. Key findings include:

1) MA-RDPG yields a notable increase in GMV, especially in store search, with moderate improvements in the main search. This is primarily because MA-RDPG directs more users from the main search to the store search rather than vice versa, benefiting the store search more substantially.

2) Comparative results with L2R+EW further validate the necessity of scenario cooperation, as optimizing the main search alone using L2R+EW negatively impacts the performance metrics of the in-store search.

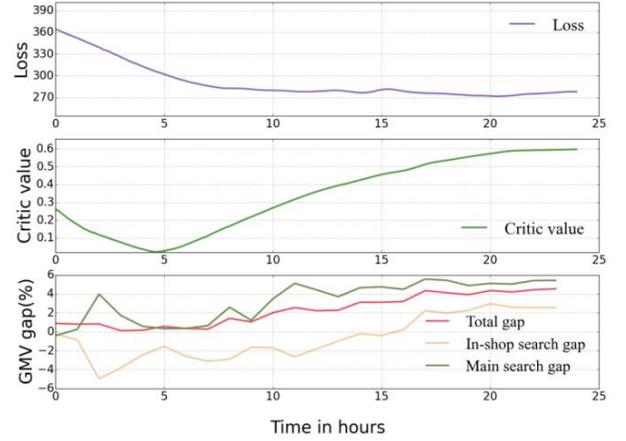

Figure 4. Top/Middle: The learning process of the critic/actor network. Bottom: The online improvement of GMV over time.

From Figure 4, we can see the change in time over which MA-RDPG improves online GMV. We can see that the algorithm improves GMV continuously and stably.

### B. Behavior Analysis

In our study, each agent displays continuous behavior, allowing us to monitor behavior changes over time across different search scenarios, as illustrated in Figure 5. We represent each behavior as a real vector and depict the average behavior across dimensions in our graphs.

The main search scenario graph shows that the most significant feature is the CTR estimation score (Action_1), affirming its relevance to ranking efficacy. Surprisingly, the second most influential feature, indicated by Action_6, is the popularity of a product's corresponding store. This feature, though not typically pivotal in Learning to Rank (L2R) models, proves crucial here for directing traffic from the main to the store search, enhancing cooperation between scenarios.

The following sub-graph describes the behavior change over time of the store search. Action_0 is the most important feature, which represents the sales volume of a product; this means that in a store, hot-selling products tend to be more likely to be sold.

Despite initial fluctuations, the action distribution stabilizes after 15 hours of training, aligning with observations from Figure 4.

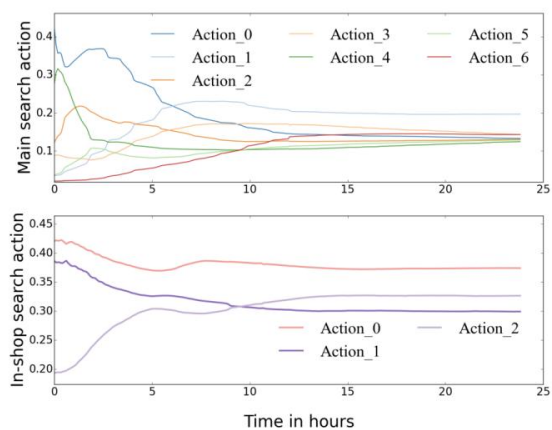

Figure 5. Changes in the average action of main search and store search.

## VI. DISCUSSION

The results highlight the potential of cooperative multi-agent systems in optimizing complex, interconnected environments like e-commerce platforms. The MA-RDPG algorithm effectively balances individual scenario goals with the overall platform objective, leading to improved global performance. The use of recurrent neural networks in the communication module enables agents to maintain contextual awareness, further enhancing decision-making accuracy.

We further analyze a typical example to illustrate how MA-RDPG makes the main search and in-store search work together. Considering that there are too many changes in online systems, we focus on analyzing some typical cases here and compare the ranking results of MA-RDPG and L2R+L2R algorithms. For example, how can the main search help in-store search to get more overall benefits. We assume such a scenario: a female user with strong purchasing power clicks on many high-priced and low-conversion products, and then searches for a keyword "dress". Obviously, MA-RDPG is more likely to return some high-priced and low-sales products in large stores, which makes it easier for users to enter the store. Compared with the L2R+L2R algorithm, MA-RDPG can sort from a more global perspective. It not only considers the current short-term clicks and transactions, but also considers potential transactions that will be made in the store.

## VII. CONCLUSION

This study presents a novel approach to multi-scenario optimization in e-commerce using the MA-RDPG algorithm. By treating multiple scenarios as cooperative agents with shared goals, we achieve significant improvements in key performance metrics. Future research can explore the application of this framework to other domains and further refine the communication mechanisms between agents.